\documentclass[letterpaper]{article} 
\usepackage{aaai2026}  
\usepackage{times}  
\usepackage{helvet}  
\usepackage{courier}  
\usepackage[hyphens]{url}  
\usepackage{graphicx} 
\usepackage{natbib}  
\usepackage{caption} 
\usepackage{algorithm}
\usepackage{algorithmic}
\usepackage{amsmath}
\usepackage{booktabs}
\usepackage{amsfonts}
\usepackage{tabularx}
\usepackage{multirow}
\usepackage{colortbl}
\usepackage{makecell}
\usepackage{newfloat}
\usepackage{listings}
\DeclareCaptionStyle{ruled}{labelfont=normalfont,labelsep=colon,strut=off} 
\floatstyle{ruled}
\newfloat{listing}{tb}{lst}{}
\floatname{listing}{Listing}
\title{Driving in Spikes: An Entropy‑Guided Object Detector for Spike Cameras}
\author{
    Ziyan Liu\equalcontrib\textsuperscript{\rm 1}
    Qi Su\equalcontrib\textsuperscript{\rm 1},
    Lulu Tang\textsuperscript{\rm 2}\thanks{Corresponding authors},
    Zhaofei Yu\textsuperscript{\rm 1}\footnotemark[2],
    Tiejun Huang\textsuperscript{\rm 1}\textsuperscript{\rm 2}
}
\affiliations{
    \textsuperscript{\rm 1}Peking University\\

    \textsuperscript{\rm 2}BAAI

}

\begin{document}

\maketitle

\begin{abstract}

Object detection in autonomous driving suffers from motion blur and saturation under fast motion and extreme lighting. Spike cameras, a type of neuromorphic sensor, offer microsecond latency and ultra-high dynamic range for object detection by using per-pixel asynchronous integrate-and-fire. However, their sparse, discrete output cannot be processed by standard image-based detectors, posing a critical challenge for end-to-end spike-stream detection. We propose EASD, an end‑to‑end spike‑camera detector with a dual‑branch design: a Temporal-Based Texture plus Feature Fusion branch for global cross‑slice semantics, and an Entropy Selective Attention branch for object‑centric details. To close the data gap, we introduce DSEC-Spike, the first driving-oriented simulated spike detection benchmark. On DSEC-Spike, EASD delivers a +5.5\% COCO mAP improvement over the best RGB+Event competitor and +11.2\% over the strongest spike baseline, establishing a new SOTA. On the real PKU-Vidar-DVS, EASD further improves by +2.2 points over the strongest spike baseline, demonstrating strong simulation-to-reality generalization. These gains indicate that spike-only detection can learn robust spatiotemporal representations and provide a reproducible paradigm and benchmark for neuromorphic perception in autonomous driving.


\end{abstract}


\section{Introduction}

Object detection has long been a fundamental research focus in computer vision, aiming to identify and localize objects accurately. Recently, image-based detection methods \cite{ref8,ref4,ref9} have achieved substantial advancements. However, object detection with conventional frame-based cameras (e.g., 30 Hz) suffers from motion blur under fast motion and saturation under extreme illumination, which limits detection performance in dynamic driving scenes. Neuromorphic vision sensors have emerged as a compelling alternative due to their high dynamic range and ultra‑fine temporal sampling, enabling robust perception where standard cameras struggle.


\begin{figure}[ht]
  \centering
  \includegraphics[width=\linewidth]{ 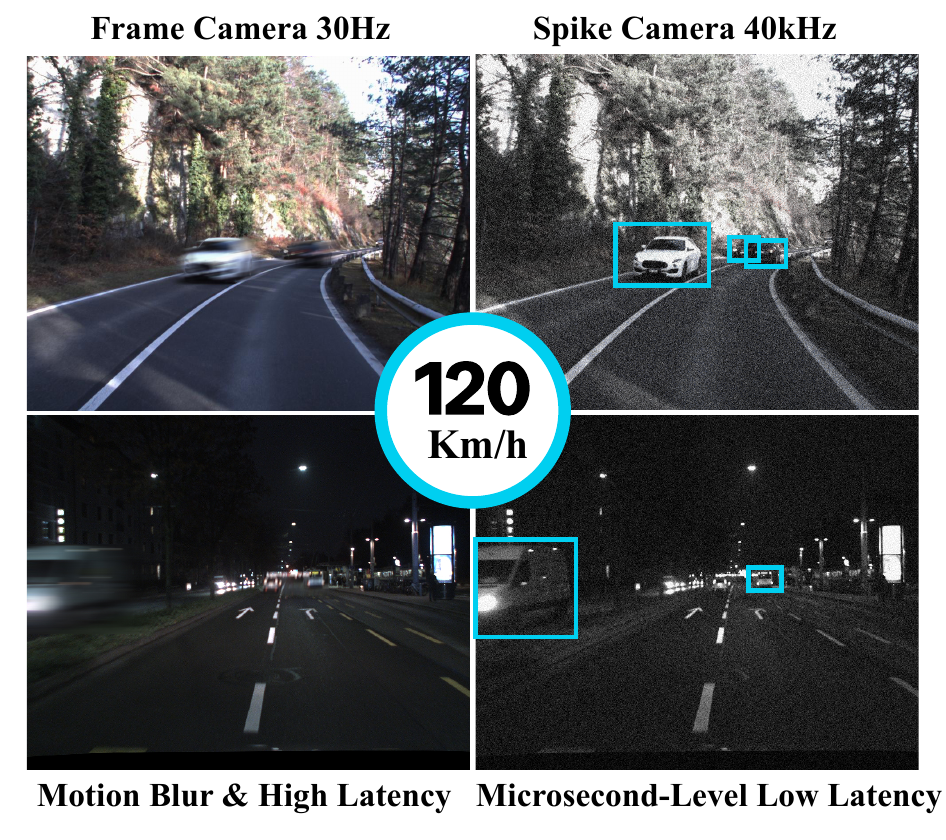}
  \caption{Illustration comparing motion blur effects in conventional cameras and spike cameras under high-speed conditions. While conventional cameras suffer from missed detections due to blur, spike cameras preserve temporal fidelity. This figure is for visualization purposes only. }
    \label{fig1}
\end{figure}


The predominant categories of neuromorphic cameras encompass event cameras and spike cameras. Event cameras \cite{ref36} asynchronously report polarity events when the log‑intensity change exceeds a threshold \cite{ref55,ref56}. \textbf{Different from event cameras}, spike cameras mimic the foveal sampling mechanism, where each pixel accumulates irradiance and emits a binary spike once a threshold is reached, then resets \cite{ref53, ref54, ref37}. This yields sparse spike streams that provide microsecond-level latency and very high dynamic range, crucially retaining static information unlike event cameras. These qualities make them well-suited for autonomous driving scenarios. However, due to the distinct working principles and data formats of spike cameras compared to event and frame-based cameras, existing image-based and event-based approaches cannot be directly applied to spike signals \cite{ref27}.

Recently, spike-based multi-object detection prototypes have demonstrated low power and ultra-low end-to-end latency, underscoring the strong potential of spike cameras for high-speed and extreme scenarios. Despite these advantages, directly performing object detection on spike streams remains underexplored. First, algorithmic challenges arise because fixed time‑window aggregation can either blur motion or miss critical spatiotemporal cues when misaligned with scene dynamics; additionally, similar reflectance between targets and background can induce comparable firing rates, complicating localization \cite{ref30}. Second, a data bottleneck persists: existing spike datasets are largely focused on texture reconstruction \cite{ref20,ref21,ref22,ref23,ref57}. Although PKU‑Vidar‑DVS \cite{ref30} provides spike+event data, its labels originate from the event stream and are misaligned with spikes due to differences in spatial resolution and viewpoints, complicating fair training and evaluation. Early spike‑stream MOT \cite{ref53} pipelines apply an SNN filter with 8‑connected search and Hungarian–Kalman association, but box‑only linkage collapses when ego‑motion also produces dense spikes. ODTSNet \cite{ref60} adds STDP‑based motion estimation and DBSCAN clustering to separate object from ego motion, yet remains hyperparameter‑sensitive with density‑dependent latency; meanwhile, an FPGA SNN‑filter implementation shows real‑time feasibility \cite{ref64}.


To address these issues, we introduce EASD, an end‑to‑end detector tailored for spike cameras. EASD adopts a dual-branch architecture to adaptively extract object-centric spatiotemporal representations in dynamic scenes: (1) an upper branch that aggregates cross‑slice global texture and semantics via a Temporal‑Based Texture Module and a Feature Fusion Module; (2) a lower branch that enhances object‑centric cues through an Entropy Selective Attention Module, adaptively focusing on regions likely to contain targets. To close the dataset gap, we construct DSEC‑Spike, a spike‑based variant of DSEC‑Detection \cite{ref15} generated by a physically consistent spike simulation pipeline \cite{ref33}, providing temporally aligned driving scenes for training and evaluation. EASD establishes a new state-of-the-art: on DSEC‑Spike, it improves over the best RGB+Event competitor by +5.5 points and over the strongest spike baseline by +11.2 points; on PKU‑Vidar‑DVS, it further gains +2.2 points over the strongest spike baseline. These consistent gains indicate that spike‑only detection can learn robust spatiotemporal representations and generalize from simulation to reality. Our work demonstrates that spike cameras, beyond their low‑power and ultra‑low‑latency sensing advantages, can support accurate, efficient multi‑object detection in difficult driving conditions.

The main contributions are summarized as follows:
\begin{itemize}
 \item[$\bullet$] We propose a dual‑branch spike‑camera detector, EASD, that couples cross‑slice global texture modeling with entropy‑guided object‑centric attention;   

  \item[$\bullet$] We construct the first simulated spike‑based detection benchmark, DSEC‑Spike, aligned with autonomous driving scenes; 

  \item[$\bullet$] SOTA detection performance with substantial gains on both DSEC‑Spike and PKU‑Vidar‑DVS, evidencing strong simulation‑to‑real generalization and practical promise for high‑speed, extreme scenarios.

\end{itemize}

\begin{figure}[t]
\centering
\includegraphics[width=0.95\linewidth]{ 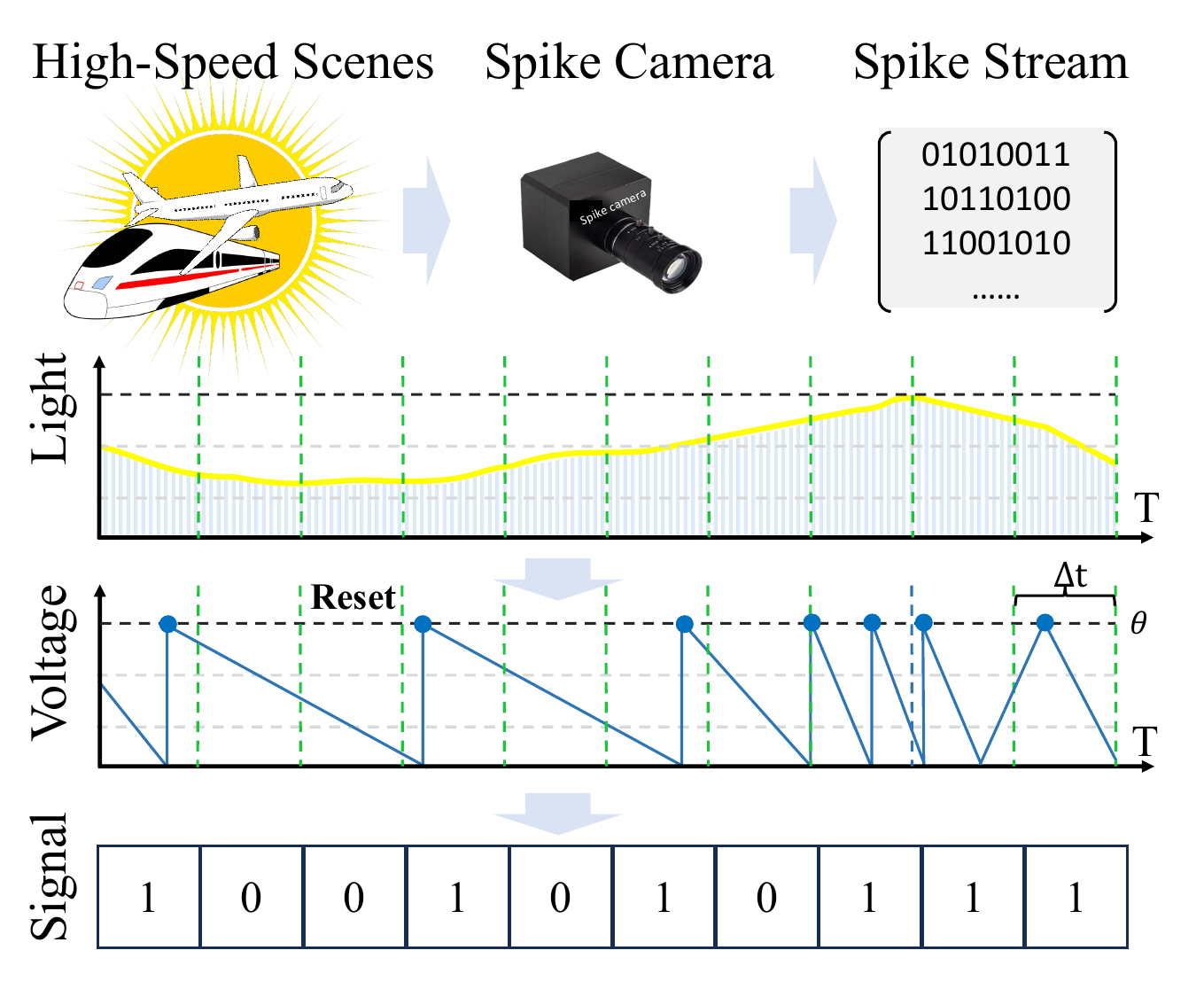} 
\caption{Illustration of the working mechanism of spike cameras, highlighting how they convert intensity changes into asynchronous spike signals.}
\label{fig2}
\end{figure}

\section{Related Work}

\noindent \textbf{Principle Of Spike Camera}. As shown in Figure~\ref{fig2},
spike cameras mimic the visual sampling mechanism of the fovea  \cite{ref53}, consisting of a photon receptor, an integrator, and a comparator. For each pixel, the receptor gathers incoming photons, the integrator converts them into a voltage, and the comparator continuously checks whether this voltage surpasses a predefined threshold $\theta$. Once exceeded, a spike is fired, and the accumulated value resets to zero. This process generates a continuous binary spike stream with varying inter-spike intervals, which can be formulated as:



\begin{equation}
\int_{k\Delta t}^{(k+1)\Delta t} \lambda I(x_n, y_n, t) \,dt \ge \theta \quad \Rightarrow \quad \text{spike}=1 ,
\label{eq1}
\end{equation}

\noindent where $I(x_{n}, y_{n},t)$ represents the light intensity at pixel position $(x_{n}, y_{n})$ and time $t$. $\lambda$ denotes photoelectric conversion rate , $\Delta t$ is the temporal resolution (eg., 25us). More details of spike cameras can be found in \cite{ref53}.

\begin{figure*}[t]
\centering
\includegraphics[width=0.9\textwidth]{ 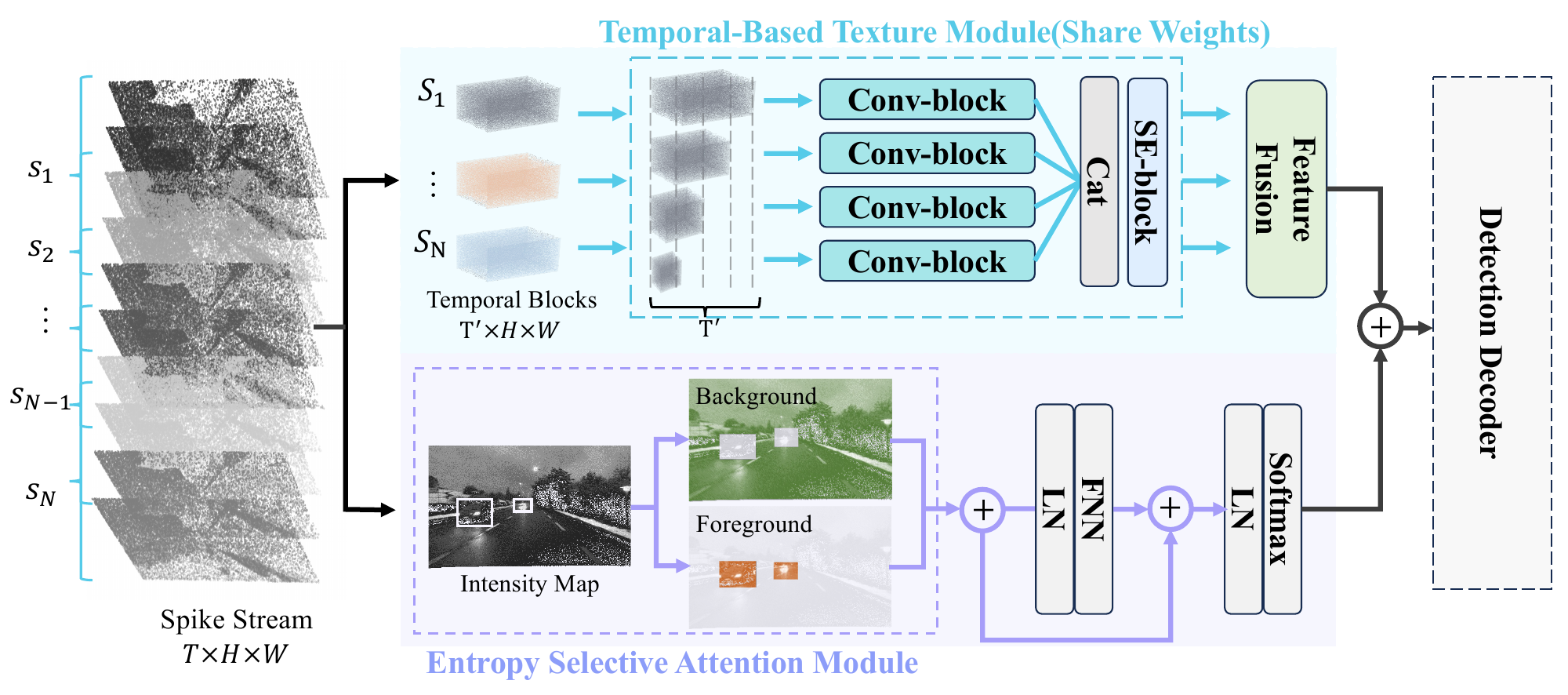} 
\caption{\textbf{Overview of EASD}. A dual-branch architecture: an upper branch that aggregates cross‑slice global texture and semantics via a Temporal‑Based Texture Module and a Feature Fusion Module; a lower branch that enhances object‑centric cues through an Entropy Selective Attention Module, adaptively focusing on regions likely to contain targets.}
\label{fig3}
\end{figure*}

\noindent \textbf{Spike-Based Vision Tasks}.
Spike-based image reconstruction has received significant attention. Early methods such as TFI and TFP \cite{ref20} utilize fixed spike intervals or spike counts for image recovery, while more recent learning-based approaches improve reconstruction through wavelet transforms \cite{ref21}, multi-scale temporal encoding \cite{ref22}, and optical flow-guided fusion \cite{ref58}. For optical flow estimation, progress has been made via self-supervised learning \cite{ref23}, motion-aware deblurring \cite{ref24}. In object recognition, Zhao et al. \cite{ref27, ref28} incorporate denoising, motion modeling, and knowledge distillation for improved performance. 

\noindent \textbf{Spike-Based Object Detection}.
Li et al. \cite{ref30} pioneer the first deep-learning-based object detection framework for spike cameras, namely Tar-Spikes+YOLO, accompanied by two hand-crafted baselines: TFI+YOLO and TFP+YOLO. The TFI/TFP+YOLO frameworks, which rely on spike signals sampled at fixed time intervals or fixed numbers, are ill-suited for diverse scene conditions: fast-moving objects require short integration intervals to mitigate motion blur and temporal aliasing, whereas low-light regions benefit from longer time intervals to ensure sufficient photon accumulation. In contrast, Tar-Spikes+YOLO leverages a learnable temporal aggregation mechanism to preserve the spatiotemporal characteristics inherent in asynchronous spike streams. Nevertheless, this approach faces limitations in object detection scenarios where spike signals of the background and targets exhibit substantial similarity due to comparable reflectance intensities. Compared with these approaches, our method, EASD, employs a dual-branch network architecture to extract object-centric features in dynamic scenes adaptively.

\section{Methodology}
\noindent \textbf{Problem Definition.} Let the spike stream from the spike camera be \( S = \left\{ (x_n, y_n, t_n): n = 1, \dots, N, \, t_n \in [0, T] \right\} \), where \( T \) is the total time length. Here, $S \in \mathbb{R}^{T \times H \times W}$ denotes a set of discrete points in three-dimensional space, with values of either 0 or 1. Our objective is to accurately detect and recognize target objects from these discrete signals. This problem can be formulated as:
\begin{align}
B_{t}=\mathbf{D_{M}}\left \{ S_{t-k}, ...,S_{t+k} \right \},
\end{align}

\noindent where $\mathbf{D_{M}}$ denotes the detection model, $S_{t-k}$ represents the spike signals at timestamp $t-k$, $t$ denotes the current timestamp, and $k$ indicates the temporal offset. $B_t = \left\{ (x_{b}, y_{b}, w_{b}, h_{b}, t) \right\}_{b = 1}^{B}$ represents a set of $B$ bounding boxes where each box is defined by its center coordinates $(x_{b}, y_{b})$, width $w_{b}$, height $h_{b}$, and the timestamp $t$.

\noindent \textbf{Network Architecture.} As illustrated in Figure~\ref{fig3}, EASD is a dual-branch, spike-based object detection framework. The upper branch adaptively extracts texture features from dynamic spike streams. Inspired by Spk2Image~\cite{spk2imgnet}, this branch segments the spike stream into overlapping temporal slices, capturing fine-grained dynamic patterns at appropriate temporal resolutions. Each slice is \textbf{individually processed by a parameter-shared Temporal-Based Texture Module} to extract local spatiotemporal features, which are then fused by a Feature Fusion Module to generate a global context.

\noindent Concurrently, the lower branch compresses the temporal dimension of the spike stream into an intensity map, enhancing object region features through an Entropy Selective Attention Module. Finally, the complementary features from both branches are combined to form an \textbf{object-centric spike representation}, which is then fed into the detection decoder, leading to enhanced object detection performance.

\section{Temporal-Based Texture Module}

As illustrated in Figure~\ref{fig3}, the Temporal-Based Texture Module employs a multi-branch convolutional architecture to extract local texture representations at multiple scales. Let \( \mathbf{S}_i \in \mathbb{R}^{T' \times H \times W} \) denote the \( i \)-th temporal block, where \( T' \) represents the block’s temporal length (as determined by Spk2Image \cite{spk2imgnet}), and \( H, W \) are the spatial dimensions. With a temporal step size \( \Delta \in \mathbb{N} \), \( \mathbf{S}_i \) is \textbf{divided into four patches} and fed into four parallel convolutional branches, the \( k \)-th branch can then be defined as:

\begin{align}
\mathbf{F}_k = \text{Conv} \left( \phi \left( \text{Conv} \left( \mathbf{S}_i^{[k\Delta : -k\Delta]} \right) \right) \right) \quad k = 0, 1, 2, 3,
\end{align}

\noindent where \( \phi(\cdot) \) is a LeakyReLU activation function, and \( \mathbf{S}_i^{[k\Delta: -k\Delta]} \) denotes temporal slicing  from index \( k\Delta \) to \( -k\Delta \). For \(k=0 \), the entire block is used without slicing. The resulting feature maps from each branch are then concatenated to form a combined representation:
\begin{align}
\mathbf{F}_{\text{cat}} = \text{Concat}(\mathbf{F}_0, \mathbf{F}_1, \mathbf{F}_2, \mathbf{F}_3),
\end{align}

\begin{figure}[t]
\centering
\includegraphics[width=0.95\linewidth]{ 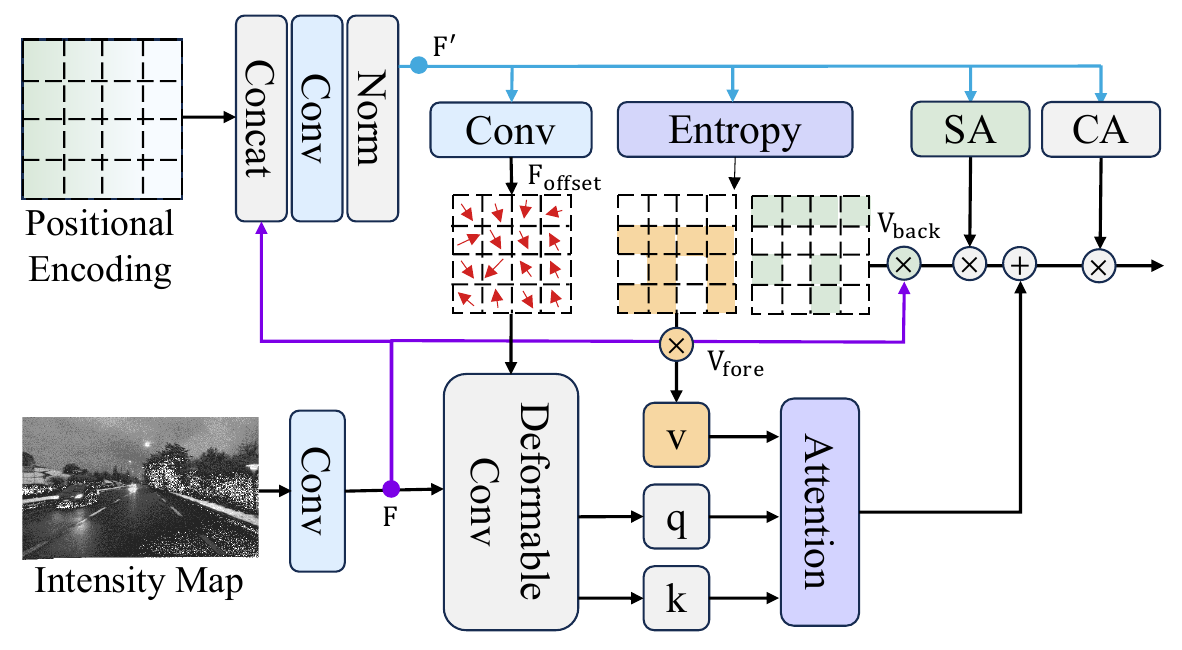} 
\caption{\textbf{Entropy Selective Attention Module}. The entropy block partitions the feature map into foreground (orange mask) and background regions (green mask). The foreground regions utilize deformable window attention for enhancement, whereas the other regions undergo simple convolution. }
\label{fig4}
\end{figure}
\vspace{-1em}


To capture multi-scale temporal dependencies, we apply the Squeeze-and-Excitation attention mechanism \cite{hu2018squeeze}, denoted as $\text{SE}(\cdot)$, to recalibrate channel-wise responses. The attended features are subsequently projected to the target output dimension:
\begin{align}
\mathbf{F}_{\text{out}} = \phi\left(\text{Conv}(\text{SE}(\mathbf{F}_{\text{cat}}))\right) \in \mathbb{R}^{C \times H \times W}.
\label{eq5}
\end{align}

\section{Feature Fusion Module}
After extracting local features from each temporal block, we obtain a set of feature maps \( \{\mathbf{F}^{(i)}_{\text{out}}\in \mathbb{R}^{C \times H \times W} \}_{i=1}^{N} \), where \( N \) denotes the number of temporal blocks. To adaptively integrate these features, we introduce a Feature Fusion Module based on channel attention. Specifically, global average pooling is applied to each feature map \(\mathbf{F}^{(i)}_{\text{out}}\), producing channel descriptors \(\mathbf{g}_i \in \mathbb{R}^{C}\). These descriptors are then concatenated to form a joint global descriptor $\mathbf{g} = [\mathbf{g}_1, \dots, \mathbf{g}_N] \in \mathbb{R}^{C \times N}$. A lightweight Multi-Layer Perceptron (MLP), with a single hidden layer and a sigmoid activation for the output, processes $\mathbf{g}$ to generate a set of normalized attention weights $\boldsymbol{\alpha} \in \mathbb{R}^{N}$, where $\boldsymbol{\alpha} = \text{MLP}(\mathbf{g})$. This allows the module to adaptively learn the importance of features from each temporal block. Finally, the fused feature $\mathbf{F}_{\text{fused}}$ is computed as a weighted sum of the original feature maps:
\begin{align}
\mathbf{F}_{\text{fused}} = \sum_{i=1}^{N} \alpha_i \cdot \mathbf{F}^{(i)}_{\text{out}}.
\end{align}
\section{Entropy Selective Attention Module}

Wang et al. \cite{ref40} propose using convolutions for simple contexts and deformable window attention for complex textures. However, their approach to region partitioning is not ideal for object detection, as textured regions may not correspond to the actual objects of interest. Inspired by this, we propose an adaptation that focuses on object-centric details. Specifically, we incorporate an entropy block to generate masks based on information entropy \cite{ref41}, which helps distinguish foreground (object-containing) regions from the background. Deformable window attention is then applied to enhance the features of the foreground, while convolutions are used to process the background. This process can be summarized as: Preprocess → Mask Generation → Foreground Feature Enhancement → Global Feature Refinement.


\noindent \textbf{Preprocess}. As shown in Figure~\ref{fig4}, the Intensity Map undergoes a $1 \times 1$ convolution to generate $\mathbf{F}$, which is concatenated with linear positional encoding and further processed through two layers to yield the intermediate feature $\mathbf{F}'$. Then, the offset map $\mathbf{F}_{\text{offset}}$ is predicted from $\mathbf{F}'$ to capture foreground displacement, which is used in the subsequent deformable window attention computation. The simplified spatial attention ($\mathbf{SA}$) and channel attention ($\mathbf{CA}$) are also derived from $\mathbf{F}'$ to support background refinement and global feature enhancement:
\begin{align}
\begin{aligned}
&\mathbf{F}_{\text{offset}}=\text{Conv}(\text{ReLU}(\text{Conv}(\mathbf{F}'))),\\
&\mathbf{SA}=\text{Mean}(\text{Conv}(\mathbf{F}'))\in \mathbb{R}^{1\times H\times W }, \\
&\mathbf{CA}=\text{Avg}(\text{Conv}(\mathbf{F}'))\in \mathbb{R}^{C\times 1\times 1  }. 
\end{aligned}
\end{align}

\begin{figure}[t]
\centering
\includegraphics[width=0.95\linewidth]{ 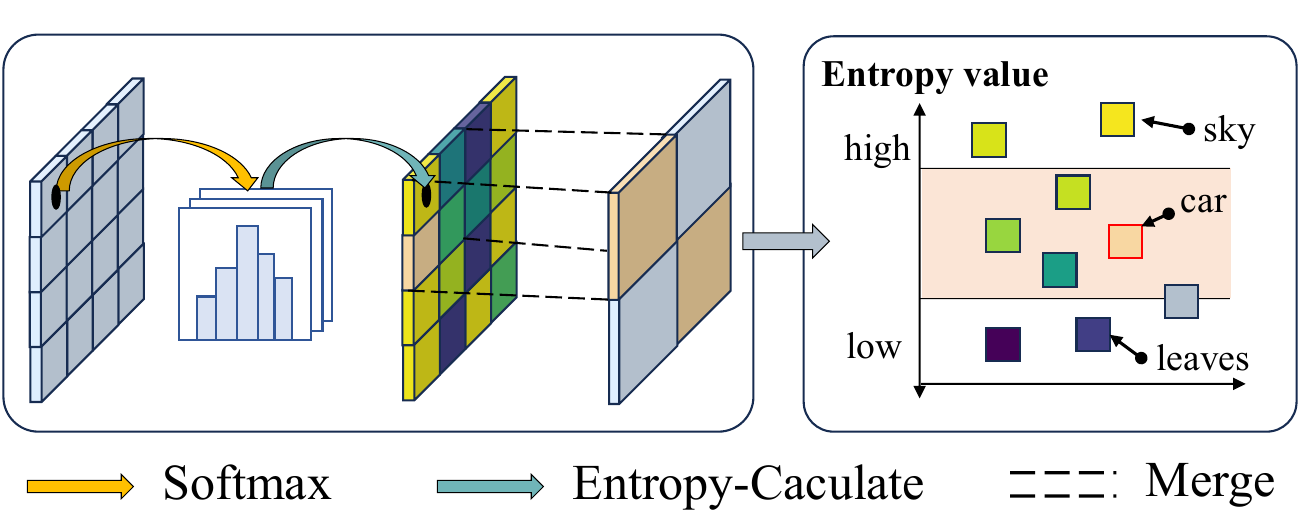} 
\caption{\textbf{Entropy Block}. The first step computes the entropy value for each window in the feature map and merges adjacent windows to calculate the average entropy. The second step selects foreground windows from the merged windows based on a specific entropy range. Squares with different colors represent different entropy levels, while the red-bordered square indicates the selected window. }

\label{fig5}
\end{figure}

\noindent \textbf{Mask Generation.} To generate masks for foreground-background separation, we introduce the entropy block. As illustrated in Figure~\ref{fig5}, the input feature $\mathbf{F}'$ is first reduced and rearranged to align with the window size $M^2$, resulting in $\mathbf{F}_v \in \mathbb{R}^{\frac{HW}{M^2} \times M^2 \times C}$. A \textit{softmax} operation is then applied within each window of $\mathbf{F}_v$ to normalize the features into probability distributions. Finally, the entropy of each window is computed as:
\begin{align}
\mathbf{E}_{n} = -\sum_{i=1}^{M^{2}} P_{n}(i) \log_2 P_{n}(i),
\end{align}

\noindent where $n \in  \left\{ 1, 2,\dots, \frac{H W}{M^2} \right\}$, $P_{n}(i)$ denotes the normalized probability of the $i$-th pixel within the $n$-th window in $\mathbf{F}_v$, and $\mathbf{E}_n$ represents the corresponding window's entropy value. To mitigate the high variance of entropy within small windows—which may hinder reliable local feature measurement—neighboring windows are merged into fixed-size blocks to compute their average entropy. The global average entropy across all windows is denoted as $\mathbf{E}_{\text{avg}}$. Merged windows with entropy values between $\mathbf{E}_{\text{avg}}$ and $\frac{1}{2}\mathbf{E}_{\text{avg}}$ are classified as foreground, while the remaining windows are considered background.\footnote{We further analyze this thresholding strategy in the experimental section.} These merged windows are then subdivided back into their original $M^2$-sized patches, and corresponding masks are generated. Let $K$ be the number of foreground windows; the foreground and background masks are denoted as $\mathbf{mask}_\text{fore} \in \mathbb{R}^{K \times M^2 \times 1}$ and $\mathbf{mask}_\text{back} \in \mathbb{R}^{(\frac{HW}{M^2} - K) \times M^2 \times 1}$, respectively.

\noindent \textbf{Foreground Feature Enhancement.}
As depicted in Figure~\ref{fig4}, following the generation of foreground--background masks, we first explicitly separate the feature map into foreground and background components using the masks:
\begin{align}
\mathbf{V}_{\text{fore}} &= \mathbf{F} \odot \mathbf{mask}_{\text{fore}}, \\
\mathbf{V}_{\text{back}} &= \mathbf{F} \odot \mathbf{mask}_{\text{back}},
\end{align}
where $\odot$ denotes element-wise multiplication, $ \mathbf{V}_{\text{fore}}$ (resp.\ $\mathbf{V}_{\text{back}}$) therefore contains non-zero activations only in the regions indicated by the corresponding mask. Subsequently, deformable window attention is applied within the foreground regions to intensively attend to object-centric representations. Specifically, the offset map $\mathbf{F}_{\text{offset}}$ guides a deformable convolution over $\mathbf{F}$, yielding a refined feature representation $\mathbf{X}$, which is reshaped to align with the dimensionality of $\mathbf{V}_{\text{fore}}$:
\begin{align}
&\mathbf{X} = f_{\text{rearrange}}\!\left( \text{DeConv}(\mathbf{F}, \mathbf{F}_{\text{offset}}) \right) \in \mathbb{R}^{K \times M^2 \times C}, \notag\\
&\mathbf{Q} = \mathbf{X}\mathbf{W}_q,\quad
\mathbf{K} = \mathbf{X}\mathbf{W}_k,\quad
\mathbf{Q},\mathbf{K}\in\mathbb{R}^{K \times M^2 \times C}, \notag\\
&\mathbf{\hat{V}}_{\text{fore}} = \mathrm{Softmax}\!\left(\frac{\mathbf{Q}\mathbf{K}^\top}{\sqrt{d}}\right)\mathbf{V}_{\text{fore}}.
\end{align}

\noindent Ultimately, deformable window attention is computed based on $\mathbf{Q}$, $\mathbf{K}$, and the mask-filtered value embeddings $\mathbf{V}_{\text{fore}}$.



\noindent \textbf{Global Feature Refinement.}
To refine spatial information while keeping the computational cost low, a lightweight spatial attention \(\mathbf{SA}\) is applied to the background regions:
\begin{align}
\hat{\mathbf{V}}_{\text{back}} = \mathbf{SA}(\mathbf{V}_{\text{back} }),
\end{align}
The enhanced foreground features \(\hat{\mathbf{V}}_{\text{fore}}\) are then fused with the refined background features \(\hat{\mathbf{V}}_{\text{back}}\).  
Finally, a channel attention \(\mathbf{CA}\) is employed to highlight discriminative object-centric information:
\begin{align}
\mathbf{V}_{\text{out}}
= \bigl(\hat{\mathbf{V}}_{\text{fore}} + \hat{\mathbf{V}}_{\text{back}}\bigr)\,\mathbf{CA}.
\end{align}

\noindent
\(\mathbf{V}_{\text{out}}\) constitutes the final output of our Entropy-Selective Attention Module.  
This adaptive, region-specific enhancement scheme cleanly separates foreground and background cues while preserving salient object details.

\section{Experiments}

\subsection{Datasets}
To evaluate the effectiveness of the proposed EASD, we conduct experiments on our newly constructed DSEC-Spike dataset and on the only publicly available neuromorphic object detection dataset, PKU-Vidar-DVS, \textbf{after its labels are corrected for the spike cameras}.

The \textbf{DSEC-Spike} dataset is constructed following the method described in SpikingSim \cite{ref33}, by generating spike data from images in DSEC-Detection \cite{ref15}. The DSEC-Spike dataset maintains \textbf{the same configuration} as the DSEC-Detection dataset. This simulation incorporates two key components: (1) the relationship between image pixel values and spike rates, and (2) noise originating from diffuse reflection and photoelectric conversion. We constructed the DSEC-Spike dataset using the default simulation parameters provided by SpikingSim \cite{ref44}. The \textbf{PKU-Vidar-DVS} dataset was collected via a dual-system setup consisting of an event camera (DVS) and a spike camera (Vidar). Key details of these two datasets are summarized in Table 1, where `Aligned' indicates whether the label is directly applicable to spike data. `Extreme Conditions' refers to scenarios with ultra-high speeds or extremely low light.

\begin{figure*}[t]
\centering
\includegraphics[width=0.9\textwidth]{ 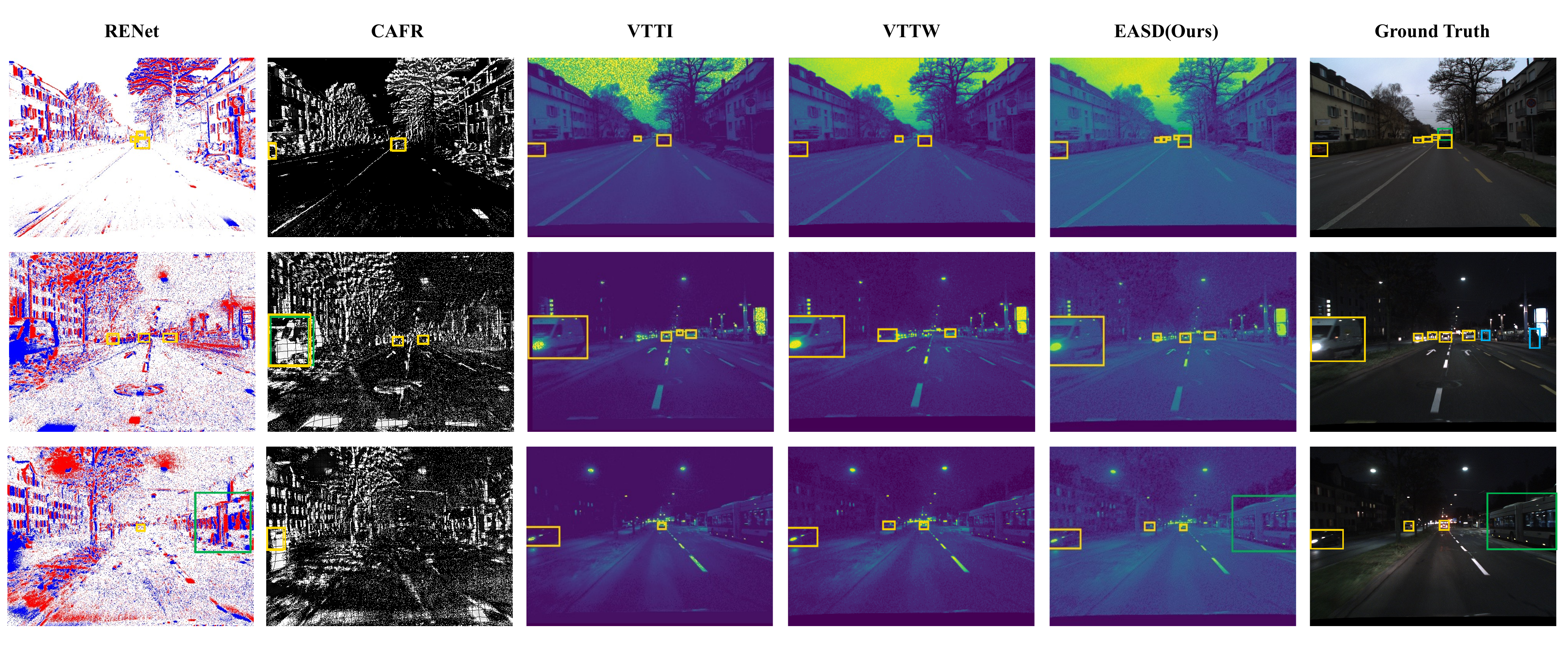} 
\caption{Qualitative results on the DSEC-Spike dataset, with event methods leveraging DSEC-Detection data. RGB ground truth from DSEC-Detection is used for visualization: yellow for cars, blue for pedestrians, and green for large vehicles.}
\label{fig6}
\end{figure*}

\begin{table*}[htbp]
  \centering
    \begin{tabularx}{\textwidth}{l l l l c X}
    \toprule
    \textbf{Dataset} & \textbf{Source} &\textbf{ Labels} & \textbf{Object} & \textbf{Aligned} & \textbf{Scenarios} \\
    \midrule
    Dsec-Spike & Simulated & 132.9k & Car, People, Large Vehicle & Yes & Driving Scenarios \\
    Pku-Vidar-DVS & Real & 215.5k & Characters, Basketball, etc. (total: 9) & No & Extreme Conditions \\
    \bottomrule
  \end{tabularx}
  \caption{Differences between the Dsec-Spike and Pku-Vidar-DVS.}
\vspace{-1em}
  \label{tab:table1}
\end{table*}

\subsection{Experimental Setup}
We adapt CSPDarkNet \cite{ref4} as the decoder and incorporate an upsampling layer to accommodate the resolution of spike signals. The proposed framework is trained using the Adam optimizer, optimizing with a standard detection loss that includes IoU loss, classification loss, and regression loss. The initial learning rate is set to 0.0001 with CosineAnnealingLR \cite{ref61} for learning rate scheduling. Training is conducted for 6 epochs on each dataset with a batch size of 8. To assess the intrinsic performance of our model, no data augmentation techniques are applied. The implementation is based on PyTorch and runs on single NVIDIA A100 GPUs. Finally, we set the overlap threshold to 0.5 and the prediction score threshold to 0.5. Performance is reported using the widely adopted COCO mAP metric.

\begin{table}[t]
  \centering
  \small
  \renewcommand{\arraystretch}{1.1}
  \setlength{\tabcolsep}{2.7mm}{
  \begin{tabular}{c|c|c|c}
    \toprule
    \textbf{Modality} & \textbf{Method} & \textbf{Venue} & \textbf{mAP@50} \\
    \midrule
    \textbf{Event} & CAFR & ECCV’24 & 12.0\% \\
    \midrule
    \multirow{5}{*}{\centering\textbf{RGB+Event}} 
        & DRFuser     & EAAI’23  & 28.1\% \\
        & CMX         & TITS’23  & 29.1\% \\
        & RENet       & ICRA’23  & 29.4\% \\
        & CAFR        & ECCV’24  & 38.0\% \\
        & FlexEvent   & arXiv’24 & \underline{47.4\%}\\
    \midrule
    \multirow{5}{*}{\centering\textbf{Spike}} 
        & VTII+YOLO    & AAAI’22  & 41.7\% \\
        & VTTW+YOLO    & AAAI’22  & 40.0\% \\
        & VTTI+RT-DETR & CVPR’24  & 34.6\% \\
        & EASD (Ours)  & --       & \textbf{52.9\%} \\
        &Improvement  & --       &\textcolor{red}{\textbf{+11.2\%}} \\
    \bottomrule
  \end{tabular}
  }
  \caption{Comparative study of state-of-the-art neuromorphic detectors on the DSEC-Detection and DSEC-Spike.}
  \label{tab:2}
  \vspace{-1em}
\end{table}

\begin{table}[htbp]
  \centering
  \setlength{\tabcolsep}{2.8mm}{
  \small
  \renewcommand{\arraystretch}{1.1}
  \begin{tabular}{c|l|l|c}
    \toprule
    \textbf{Modality} & \textbf{Method} & \textbf{Venue} & \textbf{mAP@50} \\
    \midrule
    \multirow{4}{*}{\textbf{Event}} 
        & event-image+YOLO   & IROS'18   & 32.6\% \\
        & NGA-events+YOLO    & ECCV'20   & 35.3\% \\
        & E2vid+YOLO         & CVPR'19   & 39.4\% \\
        & Tar-events+YOLO    & AAAI'22   & 38.6\% \\
    \midrule
    \multirow{6}{*}{\centering\textbf{Spike}} 
        & VTTW+YOLO          & AAAI'22   & 51.6\% \\
        & VTII+YOLO          & AAAI'22   & 55.1\% \\
        & Tar-spikes+YOLO    & AAAI'22   & \underline{57.9\%} \\
        & VTTI+RT-DETR       & CVPR'24    & 46.0\% \\
        & EASD (Ours)        &   --        & \textbf{60.1\%} \\
        & Improvement        &   --        & \textcolor{red}{\textbf{+2.2\%}} \\
    \bottomrule
  \end{tabular}
  }
    \caption{Comparative study with neuromorphic detectors detection methods on the PKU-Vidar-DVS dataset. }
  \label{tab:3}
  \vspace{-1em}
\end{table}






\subsection{Comparison with State-of-The-Art Methods}

To validate the effectiveness of our proposed method, we selected the following approaches for comparison:

\noindent\textbf{Spike-Based Detection Methods}. Given the scarcity of detection methods specifically tailored for spike cameras, we included all existing approaches: VTTI+YOLO, VTTW+YOLO, and Tar-Spikes+YOLO, as presented in \cite{ref30}. We also \textbf{integrated} VTTI \cite{ref20} with RT-DETR \cite{ref63} to establish a contemporary baseline that reflects recent advancements.

\noindent\textbf{Event-Based Detection Methods}. Spike data is often considered analogous to event data, both being forms of asynchronous neuromorphic sensor outputs. Therefore, we selected several prominent event-based models, integrating them with the YOLO detection model for comparison: Event-Image \cite{ref47}, NGA-Events \cite{ref48}, E2VID \cite{ref46}, and Tar-Events \cite{ref30}.

\noindent\textbf{Fused Event and Image-Based Detection Methods}. 
The fusion of event and RGB data currently represents a state-of-the-art approach for enhancing event-based detection performance. For this category, we chose FlexEvent \cite{ref42}, DRFuser \cite{ref49}, CMX \cite{ref44}, and RENet \cite{ref16}.

\noindent\textbf{Input Type}. Event-based baselines utilize raw event streams as input, while event-image fusion baselines integrate both event data and conventional image frames. Spike-based baselines are fed with spike data. For the experiments detailed in Table \ref{tab:2}, evaluations involving event-based and event-image fusion baselines leverage event streams and images from the DSEC-Detection dataset. Simultaneously, spike-based baselines in Table \ref{tab:2} draw their input from the DSEC-Spike dataset. For the comparisons presented in Table \ref{tab:3}, the absence of corresponding image data in the PKU-Vidar-DVS dataset necessitates the exclusion of event-image fusion detection methods from those analyses. For some methods on DSEC-Detection and PKU-Vidar-DVS, whose implementations are not publicly available, we report the results as presented in their original papers.


\begin{table}[ht]
  \centering
  \setlength{\tabcolsep}{6mm}{
  \renewcommand{\arraystretch}{1.15}
  \begin{tabular}{l|c|c}
    \toprule
    \textbf{Model} & \textbf{DSEC.} & \textbf{PKU.} \\
    \midrule
    Tar-Spikes+YOLO     & --     & 57.9\% \\
    VTTI+RT-DETR        & 34.6\% & 46.0\% \\
    Our-Upper-Branch    & 50.5\% & 56.9\% \\
    Our-Lower-Branch    & 49.6\% & 57.4\% \\
    EASD (Ours)         & 52.9\% & 60.1\% \\
    \bottomrule
  \end{tabular}
  }
  \caption{Ablations for the proposed network modules on the DSEC-Spike and PKU-Vidar-DVS datasets. The reported results are mAP@50 scores.}
  \label{tab:4}
\vspace{-1 em}
\end{table}

\begin{figure}[t]
\centering
\includegraphics[width=0.95\linewidth]{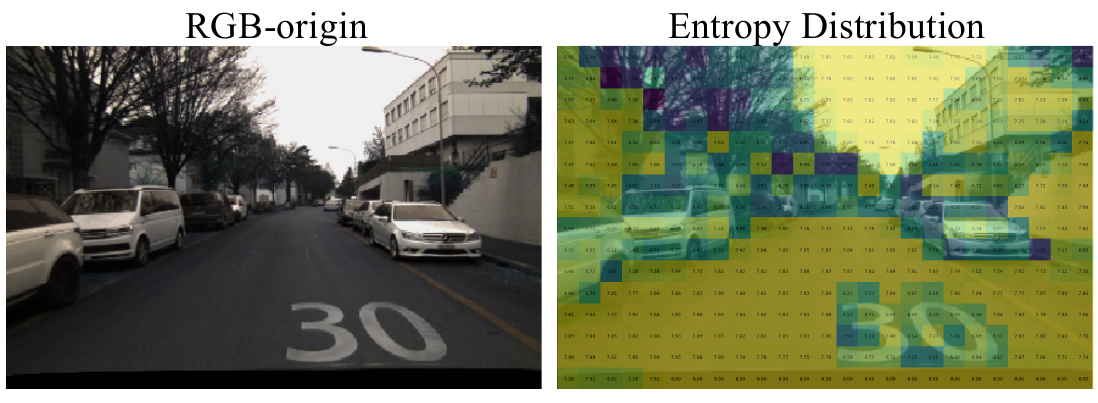} 
\caption{Feature entropy distribution in entropy blocks. RGB images are used for visualization purposes only.}

\label{fig7}

\end{figure}

\noindent\textbf{Quantitative analysis}. As presented in Table \ref{tab:2}, our approach demonstrates significant performance superiority over existing methods. Event-only methodologies often struggle with the retention of static scene information, leading to suboptimal semantic representation and consequently lower accuracy. Event-frame fusion techniques offer improved accuracy over event-only methods by leveraging complementary information from RGB frames. However, inherent frequency mismatches between conventional cameras and event cameras continue to pose challenges for effective feature fusion, particularly in highly dynamic environments. While methods such as VTTW+YOLO, VTTI+YOLO, and the VTTI+RT-DETR show promise, their coarse handling of spike streams results in considerable temporal information loss, despite the spike stream's intrinsic capability to capture both static and dynamic cues. In Table \ref{tab:3}, Tar-spikes+YOLO achieves notable detection accuracy through its effective temporal aggregation. Our method further improves upon Tar-spikes, achieving a 2.2\% performance gain. This enhancement is primarily attributed to our novel dual-branch architecture, where the upper branch adaptively captures global spatiotemporal features, while the lower branch is specifically designed to strengthen object-centric feature representations.



\noindent\textbf{Visualization}. We further provide qualitative comparisons between EASD and existing methods, as shown in Figure~\ref{fig6}. Methods like CAFR, RENet, VTTI+YOLO, and VTTW+YOLO suffer from significant missed detections, small and distant vehicles are often missed due to their limited size and subtle pixel intensity changes, while large vehicles tend to be overlooked because of the scarcity of similar samples in the training data. In contrast, our method leverages the Entropy Selective Attention Module to focus on foreground regions, achieving superior recognition performance, especially for tiny objects.

\subsection{Ablation Study and Analysis}

\begin{table}[htbp]
  \centering
    \begin{tabular}{ccccc}
    \toprule
   \textbf{Window\_Size} & \textbf{Params} & \textbf{Flops} & \textbf{DSEC.} & \textbf{PKU.} \\
    \midrule
    \midrule
    4*4   & 0.0089 M & 62M   & 52.1\% & 59.1\% \\
    8*8   & 0.0093 M & 68M   & 52.9\% & 60.1\% \\
    16*16 & 0.0129 M & 104M  & 52.2\% & 59.4\% \\
    \bottomrule
    \end{tabular}%
    \caption{Ablation study of window size on DSEC-Spike and PKU-Vidar-DVS. The reported results are mAP@50 scores.}
      \label{tab:5}%

\end{table}%

\begin{table}[htbp]
  \centering
    \begin{tabular}{ccccc}
    \toprule
   \textbf{Range} & [0, 1/2] & [0, 1] & [1/2, 1] & [1/2, 3/2] \\
    \midrule
    Dsec. & 52.0\% & 51.2\% & 52.9\% & 52.3\% \\
    PKU.  & 59.2\% & 58.4\% & 60.1\% & 59.6\% \\
    \bottomrule
    \end{tabular}%
    \caption{Ablation study of Entropy-Range on DSEC-Spike and PKU-Vidar-DVS. The results are mAP@50 scores.}
      \label{tab:6}%
\end{table}%
\noindent \textbf{Efficacy of Dual-Branch Architecture}. To validate the effectiveness of our method, we conduct ablation experiments on the DSEC-Spike and PKU-Vidar-DVS datasets, comparing our full model (EASD) and its single-branch variants with baselines Tar-Spikes+YOLO, VTII+RT-DETR. In Table~\ref{tab:4}, our Upper-Branch (UB) performs well on DSEC-Spike by leveraging temporal texture cues but struggles on PKU-Vidar-DVS due to weak object localization. In contrast, our Lower-Branch (LB) excels in the more cluttered PKU-Vidar-DVS environment by focusing on salient foreground regions, although it may filter out useful context in simpler regions. This discrepancy highlights UB's strength in texture-rich, low-noise environments versus LB's robustness in cluttered, low-visibility scenarios. EASD achieves an optimal balance through parallel processing, delivering robust and competitive performance across both datasets.

\noindent \textbf{Ablation Study on Hyperparameters}.
\textit{(1) Window Size}. As shown in Table~\ref{tab:5}, an 8x8 window size achieves the best balance between detection performance and computational overhead. Therefore, we set the default window size to 8 for all experiments in this paper. For more details, please refer to the Appendix. 
\textit{(2) Entropy Range}. We visualize the distribution of feature entropy in Figure \ref{fig7}. Table~\ref{tab:6} presents the impact of the entropy-range parameter on detection performance. Here, "Range" is quantified by the average feature entropy. Our method defaults to setting the range for selecting foreground feature windows between one-half and one times the average feature entropy. This interval is chosen because our target objects, such as vehicles, consistently exhibit feature entropy within this predefined range. Further visualizations and analysis are available in the Appendix.


\subsection{Conclusion}
We propose EASD, a dual-branch spike-based detection framework that learns object-centric representations from spike streams. We further introduce Dsec-Spike, the first simulated benchmark for spike-based object detection. Experiments on PKU-Vidar-DVS and Dsec-Spike demonstrate that EASD achieves state-of-the-art performance across diverse scenarios.

\section*{Acknowledgments}
The authors would like to thank Yajing Zheng for her valuable guidance and insightful suggestions throughout this research.

\bibliography{aaai2026}

\end{document}